\documentclass[conference]{IEEEtran}

\newif\ifnoerror
\noerrortrue

\usepackage[hyphens]{url}
\usepackage{diagbox}
\usepackage{multirow}
\usepackage[hyphenbreaks]{breakurl}
\usepackage{fancyhdr}
\usepackage{hyperref}
\fancypagestyle{firstpage}{
  \fancyhf{}
  
  \fancyfoot[L]{{\em This paper was published in the Proceedings of the 2018 IEEE 12th International 
Conference on Semantic Computing (ICSC), pages 306-307, Laguna Hills, CA, USA, January 2018. \copyright 2018 IEEE\\
Link to article abstract in IEEE Xplore: http://dx.doi.org/10.1109/ICSC.2018.00059}}
}
\urldef\spamassassinurl\url{http://csmining.org/index.php/spam-assassin-datasets.html?file=tl_files/Project_Datasets/SpamAssassin%20data/ }

\begin{document}
%
\title{Impact of Batch Size on Stopping Active Learning for Text Classification}


\IEEEoverridecommandlockouts

\author
{
\IEEEauthorblockN{Garrett Beatty\dag \thanks{\dag These students contributed equally to this paper.}}
\IEEEauthorblockA{Department of Computer Science\\
The College of New Jersey\\
Ewing, NJ 08628\\
Email: beattyg2@tcnj.edu}
\and
\IEEEauthorblockN{Ethan Kochis\dag}
\IEEEauthorblockA{Department of Computer Science\\
The College of New Jersey\\
Ewing, NJ 08628\\
Email: kochise1@tcnj.edu}
\and
\IEEEauthorblockN{Michael Bloodgood}
\IEEEauthorblockA{Department of Computer Science\\
The College of New Jersey\\
Ewing, NJ 08628\\
Email: mbloodgood@tcnj.edu}
}

\pagenumbering{gobble}

\maketitle

\thispagestyle{firstpage}

\begin{abstract}
When using active learning, smaller batch sizes are typically more efficient from a learning efficiency perspective. However, in practice due to speed and human annotator considerations, the use of larger batch sizes is necessary. While past work has shown that larger batch sizes decrease learning efficiency from a learning curve perspective, it remains an open question how batch size impacts methods for stopping active learning. We find that large batch sizes degrade the performance of a leading stopping method over and above the degradation that results from reduced learning efficiency.  We analyze this degradation and find that it can be mitigated by changing the window size parameter of how many past iterations of learning are taken into account
when making the stopping decision. We find that when using larger batch sizes, stopping methods are more effective when smaller window sizes are used. 
\end{abstract}

\section{Introduction} \label{introduction}

The use of active learning has received a lot of interest for reducing annotation costs for text classification \cite{mishler2017ICSC, bloodgood2009NAACL, schohn2000}.

Active learning sharply increases the performance of iteratively trained machine learning models by selectively determining which unlabeled samples should be annotated. The number of samples that are selected for annotation at each iteration of active learning is called the batch size. 

An important aspect of the active learning process is when to stop the active learning process. Stopping methods enable the potential benefits of active learning to be achieved in practice. Without stopping methods, the active learning process would continue until all annotations have been labeled, defeating the purpose of using active learning. Accordingly, there has been a lot of interest in the development of active learning stopping methods \cite{schohn2000, laws2008, bloodgood2009CoNLL, bloodgood2013CoNLL, vlachos2008}.

Another important aspect of the active learning process is what batch size to use.  Previous work has shown that using smaller batch sizes leads to greater learning efficiency \cite{schohn2000, brinker2003}.  There is a tension between using smaller batch sizes to optimize learning efficiency and using larger batch sizes to optimize development speed and ease of annotation. We analyze how batch size affects a leading stopping method and how stopping method parameters can be changed to optimize performance depending on the batch size. 

We evaluate the effect batch size has on active learning stopping methods for text classification. 
We use the publicly available 20Newsgroups dataset\footnote{\url{http://qwone.com/~jason/20Newsgroups/}} in our experiments.

For our base learner, we use the implementation of a Support Vector Machine from the scikit-learn Python library. For our sampling algorithm, we use the closest-to-hyperplane algorithm \cite{schohn2000}, which has been shown in recent work to compare favorably with other sampling algorithms \cite{bloodgood2018ICSC}. We use a binary bag of words representation and only consider words that show up in the dataset more than three times. We use a stop word list\footnote{Long Stopword List from \url{https://www.ranks.nl/stopwords}} to remove common English words.

For analyzing the impact of batch size on stopping methods, we use a method that will stop at the first training iteration that is within a specified percentage of the maximum achievable performance. 
We denote this method as the Oracle Method, and we will set the percentage to 99 and denote this as Oracle-99.  We set the percentage to 99 because it is typical for leading stopping methods to be able to achieve this level of performance (see Table 1 in \cite{bloodgood2009CoNLL}). Although the Oracle Method cannot be used in practice, it is useful for contextualizing the stopping results of practical stopping methods.

\section{Results} \label{results}

We considered different batch sizes in our experiments, based on percentages of the entire set of training data. The results for batch sizes corresponding to 1\%, 5\%, and 10\% of the training data for the 20Newsgroups dataset are summarized in 
Table~\ref{table:results}.

\begin{table}
\centering
\begin{tabular}{| c | c | c | c | c | c |} 
 \hline
 \diagbox[width=15em]{Stopping Method}{Batch Percent} & 1\% & 5\% & 10\%\\
 \hline
 \multirow{2}{*}{Oracle Method} & 1514.20 & 2490.40 & 3901.95  \\
 \cline{2-4} & 76.17 & 75.55 & 75.49  \\ 
 \hline
 \multirow{2}{*}{BV2009 (Window Size = 3)} & 1299.50 & 3877.10 & 6446.70  \\
 \cline{2-4} & 74.44 & 75.17 & 75.19  \\ 
 \hline
 \multirow{2}{*}{BV2009 (Window Size = 1)} & 1101.75 & 3141.30 & 5089.50  \\
 \cline{2-4} & 74.40 & 75.04 & 75.11  \\ 
 \hline
\end{tabular}
\caption{Stopping method results on 20Newsgroups for different batch sizes using various window sizes. The top number in each row shows the number of annotations at the stopping point and the bottom number shows the F-Measure at the stopping point.}
\label{table:results}
\end{table}

\subsection{Oracle Results} \label{oracleResults}

Looking at Table~\ref{table:results}, one can see that Oracle-99 needs more annotations with larger batch percents to reach approximately the same F-Measure as with smaller batch percents.

These results are consistent with past findings that learning efficiency is decreased with larger batch sizes \cite{schohn2000, brinker2003}. However, an open question is whether changing the parameters associated with actual stopping methods can make them experience less degradation in performance when larger batch sizes are used.  In particular, an important parameter of stopping methods is the window size of previous iterations to consider.  The next subsection shows how decreasing the window size parameter can help to reduce the degradation in performance that stopping methods experience with larger batch sizes. 

\subsection{Comparing BV2009 with the Oracle Method}  \label{bv2009}

We denote the stopping method published in \cite{bloodgood2009CoNLL} as BV2009.  This stopping method will stop the active learning process if the mean of the three previous kappa agreement values between consecutive models is above a threshold.  For larger batch percents, note that BV2009 stops later than the optimal Oracle Method point.

We ran BV2009 with smaller window sizes for each of our different batch sizes.  Our results are summarized for a window size of one in the row ``BV2009 (Window Size = 1)'' in Table~\ref{table:results}.  When using a window size of one, BV2009 is able to stop with a smaller number of annotations than when using a window size of three.  This is done without losing much F-Measure. The next subsection provides an explanation as to why smaller window sizes are more effective than larger window sizes when larger batch sizes are used.

\subsection{BV2009 Window Size Discussion} \label{bv2009WindowSizeDiscussion}

We set $n$ to be the window size that the user has defined.  Kappa is an agreement metric between two models.  Therefore, BV2009 needs $n+1$ models to be generated before it begins to check if the average is above the threshold.  This does not necessarily mean it stops after $n+1$ models have been generated.  Rather, it represents the first point in the active learning process at which BV2009 even has a chance to stop.

When using larger batch percents, fewer models are generated than when using smaller batch percents.  This gives any stopping method less points to test whether or not to stop.  We also note that kappa agreement scores are generally low between the first few models trained.  This, combined with fewer points to stop at, causes BV2009 to stop somewhat sub-optimally when using very large batch percents.  Usage of very large batch sizes, such as 10\% of the data, is not common so sub-optimal performance of stopping methods in those situations is not a major problem. 


\section{Conclusion} \label{conclusion}

Active learning has the potential to significantly reduce annotation costs. Two important considerations in the active learning process are when to stop the iterative process of asking for more labeled data and how large of a batch size to use when asking for additional labels during each iteration. We found that stopping methods degrade in performance when larger batch sizes are used. The degradation in performance is larger than the amount that can be explained due to the degradation in learning efficiency that results from using larger batch sizes. An important parameter used by stopping methods is what window size of earlier iterations to consider in making the stopping decision. Our results indicate that making the window size smaller helps to mitigate the degradation in stopping method performance that occurs with larger batch sizes.

\section*{Acknowledgment} \label{acknowledgment}

This work was supported in part by The College of New Jersey Support of Scholarly Activities (SOSA) program, by The College of New Jersey Mentored Undergraduate Summer Experience (MUSE) program, and by usage of The College of New Jersey High Performance Computing System.

\bibliographystyle{IEEEtran}
\bibliography{paper}

\end{document}